\documentclass[nohyperref]{article}

\usepackage{microtype}
\usepackage{graphicx}
\usepackage{booktabs} 
\usepackage{amsmath}
\usepackage{textcomp}
\usepackage[dvipsnames,table]{xcolor}

\definecolor{msMagenta}{RGB}{204,0,153}
\definecolor{msCyan}{RGB}{0,190,230}
\definecolor{msOrange}{RGB}{255,140,0}
\definecolor{msGreen}{RGB}{0,153,76}

\usepackage{pgfplots}
\usepackage{pgfplotstable}
\pgfplotsset{compat=1.18}

\usepackage{tikz}
\usetikzlibrary{arrows.meta, positioning, shapes.geometric, shapes.misc, fit, calc}

\usepackage{hyperref}

\usepackage[accepted]{icml2022}

\icmltitlerunning{Moonshine v2}

\begin{document}

\twocolumn[
  \icmltitle{Moonshine v2: Ergodic Streaming Encoder ASR for Latency-Critical Speech Applications}

  \begin{icmlauthorlist}
    \icmlauthor{Manjunath Kudlur}{}
    \icmlauthor{Evan King}{}
    \icmlauthor{James Wang}{}
    \icmlauthor{Pete Warden}{}
  \end{icmlauthorlist}

  \begin{center}
    \href{https://moonshine.ai}{Moonshine AI}
  \end{center}

  \icmlcorrespondingauthor{Manjunath Kudlur}{keveman@moonshine.ai}

  \icmlkeywords{Keywords}

  \vskip 0.3in
]

\printAffiliationsAndNotice{}

\begin{abstract}
Latency-critical speech applications—including live transcription, voice
commands, and real-time translation—demand low time-to-first-token (TTFT) and
high transcription accuracy, particularly on resource-constrained edge devices.
Full-attention Transformer encoders remain a strong accuracy baseline for
automatic speech recognition (ASR) because every frame can directly attend to
every other frame, which resolves otherwise locally ambiguous acoustics using
distant lexical context. However, this global dependency incurs quadratic
complexity in sequence length, inducing an inherent
\,``encode-the-whole-utterance''\, latency profile. For streaming use cases,
this causes TTFT to grow linearly with utterance length as the encoder must
process the entire prefix before any decoder token can be emitted. To better
meet the needs of on-device, streaming ASR use cases we introduce Moonshine v2,
an ergodic streaming-encoder ASR model that employs sliding-window
self-attention to achieve bounded, low-latency inference while preserving strong
local context. Our models achieve state of the art word error rates across
standard benchmarks, attaining accuracy on-par with models 6x their size while
running significantly faster. These results demonstrate that carefully designed local
attention is competitive with the accuracy of full attention at a fraction of
the size and latency cost, opening new possibilities for interactive speech
interfaces on edge devices.
\end{abstract}

\section{Introduction}

Modern automatic speech recognition (ASR) systems are separated into two
deployment paradigms: cloud-based models that leverage server-scale compute and
edge models that run locally on resource-constrained devices. While cloud ASR
can achieve excellent accuracy by utilizing large models and extensive
computational resources, edge ASR is essential for applications where network
connectivity is unreliable or unavailable, such as offline voice assistants,
medical dictation in remote settings, real-time captioning for accessibility,
and privacy-sensitive voice commands on mobile devices. Edge deployment also
eliminates network round-trip latency and reduces privacy concerns by keeping
audio data on-device.

In edge use cases, latency and transcription quality are the two key---and often
competing---constraints. Achieving human-perceivable real-time performance
requires minimizing time-to-first-token (TTFT) and maintaining low per-token
latency, while simultaneously delivering word error rates (WERs) competitive
with cloud-based alternatives. Balancing these competing objectives on devices
with limited memory, compute, and power budgets remains a central challenge in
practical ASR deployment.

Existing edge ASR models leverage a full-attention encoder architecture, which
allows every frame to directly attend to every other frame in a sequence of
speech audio. This enables powerful contextual disambiguation as it resolves
locally ambiguous acoustics using distant lexical information that occurs
earlier or later in a chunk of speech audio. However, full attention also
introduces quadratic complexity in sequence length and imposes an inherent
``encode-the-whole-utterance'' latency profile: in streaming scenarios, the
encoder must process the entire prefix (or wait for the complete utterance)
before decoder tokens can be emitted, resulting in high TTFT that scales
linearly with utterance length. In practical applications, this reduces system
responsiveness and limits interactivity.

In this paper, we introduce Moonshine v2, a family of ergodic streaming encoder
ASR models designed specifically for latency-critical edge applications.
Moonshine v2 models employ sliding-window attention in a position-free encoder
to enable low-latency streaming inference while maintaining state-of-the-art
accuracy on standard benchmarks. We train three variants of increasing size---
tiny, small, and medium---and show that the models achieve transcription quality and
speed on-par with models 6x their size while running significantly faster (i.e., Whisper Large v3). 
We release the models under a permissive license, encouraging community adoption for on-device, 
latency-critical ASR applications.

The paper is structured as follows. Section~\ref{sec:motivation} analyzes the
latency-accuracy trade-offs inherent in full-attention encoders and motivates
our sliding window approach. Section~\ref{sec:approach} details the Moonshine v2
architecture, including the audio preprocessor, sliding-window encoder, adapter,
and decoder components. Section~\ref{sec:evaluation} presents our experimental
setup and benchmark results across standard ASR datasets. Finally,
Section~\ref{sec:conclusion} discusses implications and future directions for
ergodic streaming ASR.

\begin{figure}[t]
  \centering
  \begin{tikzpicture}
    \begin{axis}[
      width=\columnwidth,
      height=0.65\columnwidth,
      xlabel={Audio processed (s)},
      ylabel={TTFT (ms)},
      xmin=0, xmax=10,
      ymin=0, ymax=700,
      grid=both,
      legend columns=2,
      legend style={font=\scriptsize, draw=none},
      legend to name=ttftlegend,
    ]

    %

    \addplot[very thick, color=msMagenta] expression[domain=0:10, samples=101]{(30*x + 0.09216*x^2)/0.1};
    \addlegendentry{0.1 TOPS}

    \addplot[very thick, color=msCyan] expression[domain=0:10, samples=101]{(30*x + 0.09216*x^2)/0.5};
    \addlegendentry{0.5 TOPS}

    \addplot[very thick, color=msOrange] expression[domain=0:10, samples=101]{(30*x + 0.09216*x^2)/1};
    \addlegendentry{1 TOPS}

    \addplot[msMagenta, very thick, dashdotdotted] coordinates {(0,120.15) (10,120.15)};
    \addlegendentry{0.1 TOPS (sliding window, $w{=}20$)}

    \addplot[black!70, dashed, thick] coordinates {(0,250) (10,250)};
    \addlegendentry{250 ms voice-delay limit}

    \end{axis}
  \end{tikzpicture}

  \vspace{-0.3em}
  \ref{ttftlegend}

  \caption{Illustrative time-to-first-token (TTFT) for a \emph{full-attention}
  encoder as a function of audio length, for processors with different peak
  throughput (TOPS). The estimate includes both a linear non-attention term and
  a quadratic self-attention mixing term. The dotted horizontal line shows a 0.1~TOPS 
  sliding-window encoder with $w=20$ frames.
  The dashed line indicates a 250~ms one-way delay limit often used as a
  practical upper bound for acceptable interactive voice in private
  networks~\citep{cisco_delay_details}.}
  \label{fig:ttft_full_attention}
\end{figure}
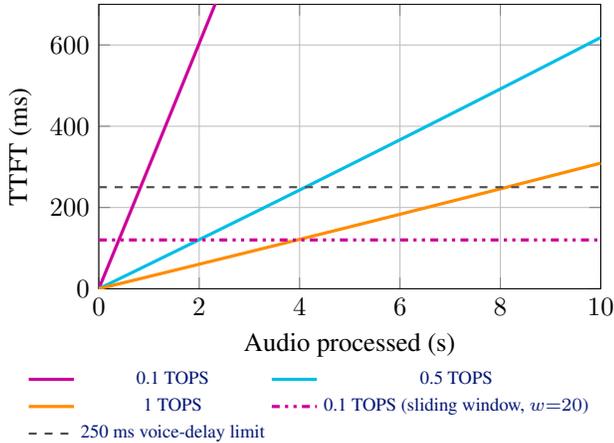

\section{Motivation}
\label{sec:motivation}
This section motivates the need for low-latency, streaming-friendly encoder
architectures in ASR, highlighting the trade-offs between recognition accuracy
and time-to-first-token (TTFT) latency in current models.

\subsection{Full-attention encoders: accurate, but latency-heavy}
Many high-accuracy ASR systems rely on encoder architectures that use full
self-attention over the entire input sequence. For example,
Whisper~\citep{radford2022robustspeechrecognitionlargescale} uses a Transformer
encoder with global attention, and NVIDIA's Parakeet models build on
FastConformer-style encoders~\citep{rekesh2023fastconformer}.

Full attention helps accuracy because it lets each frame incorporate evidence
from any other frame, enabling global disambiguation (e.g., long-range
coarticulation, speaker/style consistency, and resolving locally ambiguous
acoustics using distant lexical context). This ability to integrate long-range
context is one reason these models achieve strong recognition accuracy. However,
this same global dependency that enables superior accuracy also creates a
fundamental latency bottleneck for streaming applications.

\paragraph{Time-to-first-token (TTFT).}
For latency-critical ASR, a key metric is TTFT: the wall-clock time from audio
arrival to the first emitted text token. With a full-attention encoder, TTFT
grows with the amount of audio that must be encoded before decoding can start.
Moreover, even with a fixed model size, the attention mixing work grows
quadratically with sequence length.

Figure~\ref{fig:ttft_full_attention} illustrates this effect for a
100M-parameter encoder processing 50~Hz features (Whisper-style). We estimate
encoder compute as $\mathrm{ops}_{\mathrm{total}}(N) = 6PT + 4dLT^2$ with
$T=50N$ frames, and convert operations to time assuming a peak throughput of $X$
TOPS (i.e., $X\cdot10^{12}$ ops/s). The plotted curves show the resulting TTFT
(ms) versus audio duration for several hardware budgets. We also include a
constant TTFT line for sliding-window attention at 0.1~TOPS using
$\mathrm{ops}_{\mathrm{total}}(N) = 6PT + 4dLTw$ with $w=20$ frames (matching the
Moonshine v2 streaming lookback+lookahead window).

We plot only 0.1--1~TOPS because our focus is edge deployment (phones and
smaller devices), where achievable throughput is often in the 10s--100s of GOPS.
A simple sanity check is
\emph{peak MAC/cycle} $\approx$ (instr/cycle)$\times$(MAC/instr), e.g., an Arm Cortex-A55 might reach $\approx 16$~MAC/cycle; at 2.31~GHz this is $\approx 37$~GMAC/s ($\approx 74$~GOPS). Even when edge devices advertise multi-10s of TOPS, sustaining 1~TOPS in practice is difficult due to memory bandwidth and thermals, so we focus on the 0.1--1~TOPS regime. The horizontal line at 250~ms marks a commonly used one-way delay limit for acceptable interactive voice in private networks~\citep{cisco_delay_details}.

A key takeaway is that even a very strong edge-class budget of 500~GOPS
($\approx 0.5$~TOPS) crosses the 250~ms threshold at roughly 4.1~s of audio in
this model, making ``responsive'' first-token latency impractical for longer
utterances without streaming.

For sliding-window attention, we show only the 0.1~TOPS line because it already
falls below the 250~ms voice-delay limit; higher-throughput hardware would
reduce the line further.

\subsection{Sliding-window attention encoders: streaming-friendly latency}
A natural way to reduce TTFT is to replace full self-attention with
\emph{sliding-window} self-attention, where each frame attends only to a bounded
local neighborhood. With a fixed window size $w$, the attention mixing cost
becomes linear in sequence length ($\mathcal{O}(Tw)$ rather than
$\mathcal{O}(T^2)$), and—crucially for streaming—the encoder can emit usable
representations incrementally as soon as the required local context has arrived.

In a causal sliding-window encoder, the representation at time $t$ depends only
on past frames, so it can be produced immediately without waiting for future
audio. If a small right context is used (lookahead), the algorithmic latency is
bounded by $w_\text{right}$ frames (e.g., $w_\text{right}\times20\,\mathrm{ms}$
at 50~Hz). This bounded, constant lookahead makes latency predictable and
largely independent of utterance duration, enabling responsive partial
hypotheses for live transcription.

\begin{figure*}[t]
    \centering
    \includegraphics[width=\textwidth]{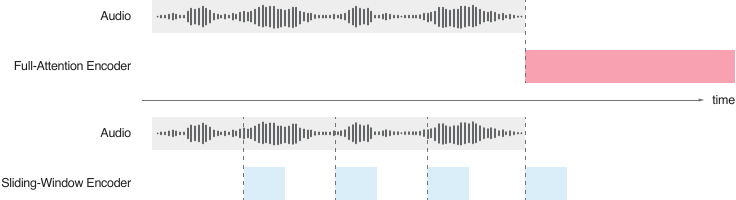}
    \caption{Conceptual TTFT timelines. In full attention, encoding begins after the entire audio has arrived. With sliding-window attention, encoding proceeds incrementally and overlaps with audio capture, so the remaining work after the last chunk is smaller.}
    \label{fig:latency-timelines}
\end{figure*}

\begin{figure}[t]
    \centering
    \includegraphics[width=.85\columnwidth]{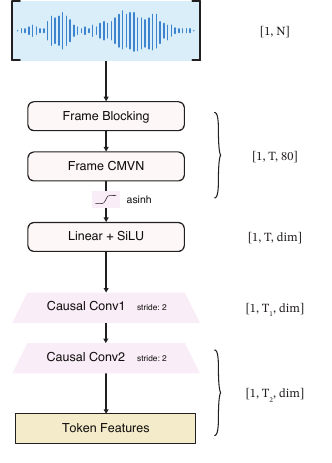}
    \caption{Audio preprocessor overview with tensor shapes, single-example. Dimensions $T = \lfloor \frac{N}{80} \rfloor$, $T_1 = \lceil \frac{T}{2} \rceil$, and $T_2 = \lceil \frac{T_1}{2} \rceil$}
    \label{fig:audio-preprocessor}
\end{figure}

\section{Approach}
\label{sec:approach}
Moonshine v2 consists of four high-level stages: an audio preprocessor, a
streaming encoder, an adapter, and a decoder. We start by detailing the audio
preprocessor.

\subsection{Audio preprocessor}
Our audio preprocessor is intentionally lightweight: it converts raw audio to a
50~Hz feature sequence (matching Whisper's feature rate) using simple operations
with no right context. Many of the frontend choices were informed guesses and
engineering intuition rather than a comprehensive ablation study; a full sweep
over alternative frontends is cost-prohibitive for us and out of scope for this
paper.

The original Moonshine model~\citep{jeffries2024moonshinespeechrecognitionlive}
used a full-attention encoder and a different frontend with an effective feature
rate of 41.6~Hz. In Moonshine v2 we standardize on 50~Hz features to align with
Whisper~\citep{radford2022robustspeechrecognitionlargescale} and to simplify
comparisons.

Specifically, the frontend processes audio by segmenting it into non-overlapping
80-sample windows (equivalent to 5~ms at 16~kHz), performing per-frame cepstral
mean and variance normalization (CMVN)~\citep{acero1995augmentedcepstralnorm},
and applying an $\operatorname{asinh}$ nonlinearity. The $\operatorname{asinh}$
function, like $\tanh$, is smooth and nearly linear around zero, but it
increases logarithmically for large values rather than saturating, which we
found balances compression and dynamic range effectively. Finally, two causal
stride-2 convolutions reduce the frame rate by approximately a factor of four,
yielding about 50 feature frames per second.

\subsection{Encoder}
The encoder is a standard Transformer stack with sliding-window self-attention.
Each layer attends to a fixed number of past frames (left context) and,
optionally, a small number of future frames (right context). We denote the
attention window as $(w_\text{left}, w_\text{right})$ in frames.

\paragraph{No positional embeddings (ergodic encoder).}
We do not use any absolute or relative positional embeddings in the encoder. As
a result, encoder computations are translation-invariant in time: for any local
window, the same function is applied regardless of where that window occurs in
the utterance. Informally, the encoder is \emph{ergodic} in the sense that it
has no explicit notion of absolute position; it can only infer structure from
the content of the local context provided by sliding-window attention.

In Moonshine v2, we use $(16,4)$ for the first two and last two encoder layers,
and $(16,0)$ for all intermediate layers. Since each encoder input frame
corresponds to 20~ms of audio (50~Hz), a right window of $w_\text{right}=4$
implies an algorithmic lookahead of $4\times20\,\mathrm{ms}=80\,\mathrm{ms}$: to
produce the representation at time step $t$ for layers with lookahead, the model
may use information up to frame $t+4$, i.e., up to 80~ms of future audio.

Layers with $(16,0)$ are strictly causal: their output at time $t$ depends only
on frames $\leq t$ (plus whatever future information has already been mixed into
the current frame by earlier lookahead layers). Overall, this design keeps
encoder lookahead bounded and small while still allowing limited future context
near the bottom and top of the stack.

\paragraph{Provisional vs. finalized encoder states.}
We note that the right-context layers also imply that, in steady state, a
\emph{finalized} representation for time step $t$ cannot be produced until
additional future audio has arrived. In our setting, a conservative bound is 16
frames of extra audio, i.e., $16\times20\,\mathrm{ms}=320\,\mathrm{ms}$ of
future context.

For applications such as live caption display, we can still decode from
\emph{provisional} (not-yet-finalized) encoder states: the newest suffix may be
less accurate, but as more audio arrives the provisional states are overwritten
by finalized ones and the displayed transcription naturally improves.

\subsection{Adapter}
The adapter bridges the ergodic encoder and the decoder. It adds a learned
positional embedding to the encoder outputs, and (when needed) applies a linear
projection so that the representation dimension matches the decoder dimension.
In other words, the encoder remains position-free, while the decoder receives
position-aware inputs.

\subsection{Decoder}
The decoder is a standard causal Transformer with rotary positional embeddings
(RoPE) in each layer~\citep{su2023roformerenhancedtransformerrotary}. It
autoregressively generates text tokens and cross-attends to the adapter
features.

While our ergodic streaming encoder makes the first usable features available
quickly (and thus TTFT can be very low), the decoder remains autoregressive:
generating a long transcript still requires a token-by-token loop, which adds
latency to the full output.

A fully ergodic, infinite-streaming alternative would be to predict directly
from encoder features using a linear classifier trained with
CTC~\citep{graves2006ctc}, or to use a monotonic transducer objective such as
RNN-T~\citep{graves2012rnnt} or Token-and-Duration Transducer
(TDT)~\citep{xu2023tdt}. Parakeet-class models follow this general direction
(CTC/RNN-T/TDT-style training and decoding) and shift much of the modeling
capacity into a larger encoder~\citep{rekesh2023fastconformer}. Marrying these
objectives with our position-free (ergodic) encoder is a promising direction
that we leave to future work.

\section{Evaluation \& Results}
\label{sec:evaluation}
We trained three model sizes (Table~\ref{tab:model-sizes}) and evaluate them on
standard ASR benchmarks and latency-sensitive streaming scenarios.

\begin{table*}[t]
    \centering
    \small
    \setlength{\tabcolsep}{4pt}
    \begin{tabular}{l c c c r r r r r}
    \toprule
    & \multicolumn{3}{c}{Architecture} & \multicolumn{5}{c}{Params (M)} \\
    \cmidrule(lr){2-4} \cmidrule(lr){5-9}
    Model & Enc dim & Dec dim & Layers (Enc/Dec) & Pre & Enc & Adap & Dec & Total \\
    \midrule
    Tiny   & 320 & 320 & 6/6   & 2.08 & 7.39 & 1.31 & 22.80 & 33.57 \\
    Small & 620 & 512 & 10/10 & 7.74 & 43.49 & 2.86 & 69.27 & 123.36 \\
    Medium & 768 & 640 & 14/14 & 11.86 & 93.66 & 3.64 & 135.77 & 244.93 \\
    \bottomrule
    \end{tabular}
    \caption{Moonshine v2 model architecture sizes and parameter breakdown by block.}
    \label{tab:model-sizes}
\end{table*}

\noindent\textbf{Note on parameter distribution.} The decoder has substantially more parameters than the encoder, largely because each decoder layer includes additional cross-attention projection matrices (in addition to self-attention), and because our decoder uses SwiGLU feed-forward blocks while the encoder does not.

\subsection{Experimental setup}
\subsubsection{Training.}
\paragraph{Data.} We use the same data sources and preprocessing pipeline as in the original
Moonshine work~\citep{jeffries2024moonshinespeechrecognitionlive} (see their
Section~3.2, \emph{Training data collection \& preprocessing}). Relative to that
setup (\,$\approx$200K hours total), we add an additional 100K hours of
internally prepared speech data, for a total of roughly 300K hours.

\paragraph{Tokenizer \& optimization. }We use the same tokenizer as in the original Moonshine
work~\citep{jeffries2024moonshinespeechrecognitionlive}. We also use the same
optimizer (Schedule-Free;~\citep{defazio2024roadscheduled}) with a starting
learning rate of $2\times10^{-3}$. Training was run for 400K steps with an
effective batch size of 512 on a cluster of 8 NVIDIA H100 GPUs.

\subsubsection{Implementation.}
We evaluate accuracy using the implementation in the Transformers
library~\citep{wolf2020transformers}. Note that this code path does not yet
perform fully efficient streaming; it relies on the flash-attention backend's
sliding-window attention when available. We measure latency separately
using our own library implementation of Moonshine in C++, which leverages the ONNX
runtime on CPU.~\footnote{https://github.com/moonshine-ai/moonshine}

\subsubsection{Benchmarks.}
We evalute the performance of Moonshine v2 using the following benchmarks.

\paragraph{Word error rate (WER).} We evaluate the WER of Moonshine v2 variants against similarly-sized models on the Open ASR leaderboard~\citep{srivastav2025open}.

\paragraph{Time-to-first-token (TTFT).} We empirically establish the latency differences between full and sliding window attention by comparing TTFT measurements of the original Moonshine~\citep{jeffries2024moonshinespeechrecognitionlive} full attention encoder to the Moonshine v2 sliding window encoder.

\paragraph{Response latency.} We perform empirical latency evaluations between Moonshine v2, the original Moonshine models, and the OpenAI Whisper models~\citep{radford2022robustspeechrecognitionlargescale} (as implemented in faster-whisper~\footnote{https://github.com/SYSTRAN/faster-whisper}). ASR models like Whisper were originally intended for offline processing scenarios, where the overall throughput of the system is important. Since our use case targets online processing applications (e.g., live captioning), we measure the real-time \emph{response latency} rather than throughput. We define this as the amount of time taken between detecting the end of a speech segment in an audio stream (via a voice activity detection (VAD) model) and the transcript text being returned. This is representative of, e.g,. the amount of time an embedded voice command system might take to detect a command after a typical utterance. We compare against Whisper because it has been adopted in embedded applications (via, e.g., whisper.cpp) despite its original design intent as an offline model, and against the original Moonshine models since they were designed for real-time use cases.

\paragraph{Compute cost.} We empirically measure compute cost by totalling the duration of the audio processing times for each model, and then expressing that as a percentage of the total audio duration. This is the inverse of the commonly used real-time factor (RTF) metric, but it reflects the compute load required for a real-time application.

We run empirical evaluations on an Apple M3.

\subsection{Results}
\begin{figure*}[t]
  \centering
  \small
  \pgfplotstableread[col sep=comma]{
params,wer
2500,5.63
8000,5.74
2000,6.00
5600,6.02
600,6.05
600,6.32
1000,6.35
2600,6.40
1000,6.50
245,6.65
1550,6.67
1100,7.01
3000,7.05
1100,7.12
180,7.12
1000,7.15
756,7.21
1000,7.23
1100,7.40
1000,7.42
1000,7.43
1550,7.44
110,7.49
600,7.50
756,7.52
600,7.69
809,7.77
809,7.83
1550,7.83
123,7.84
756,7.92
1550,7.94
769,8.09
1000,8.12
1000,8.25
8000,8.52
166,8.57
244,8.59
394,8.77
61,9.99
74,10.32
19,10.61
34,12.01
27,12.65
39,12.81
} \openasravgall

  \pgfplotstableread[col sep=comma]{
params,wer,label
34,12.01,Moonshine v2 Tiny
123,7.84,Moonshine v2 Small
245,6.65,Moonshine v2 Medium
} \moonshinevTwo

  \pgfplotstableread[col sep=comma]{
params,wer
1550,7.44
809,7.83
1550,7.83
1550,7.94
769,8.09
244,8.59
74,10.32
39,12.81
} \openaiModels

  \pgfplotstableread[col sep=comma]{
params,wer
2500,5.63
600,6.05
600,6.32
1000,6.35
1000,6.50
1100,7.01
1100,7.12
180,7.12
1000,7.15
1100,7.40
110,7.49
600,7.50
600,7.69
} \nvidiaModels

  \pgfplotstableread[col sep=comma]{
params,acc
34,87.99
123,92.16
245,93.35
} \moonshineFrontier

  \pgfplotstableread[col sep=comma]{
params,acc
39,87.19
74,89.68
244,91.41
769,91.91
809,92.17
1550,92.56
} \openaiFrontier

  \pgfplotstableread[col sep=comma]{
params,acc
110,92.51
180,92.88
600,93.95
2500,94.37
} \nvidiaFrontier

  \begin{tikzpicture}
    \begin{axis}[
      width=\textwidth,
      height=0.45\textwidth,
      xmode=log,
      log basis x=10,
      xmin=10,
      xmax=10000,
      ymin=86,
      ymax=95,
      xlabel={Parameters (M, log scale)},
      ylabel={Accuracy (100 - WER, \%)},
      grid=both,
      major grid style={line width=0.2pt, draw=gray!30},
      minor grid style={line width=0.1pt, draw=gray!15},
      tick align=outside,
      tick style={line width=0.2pt},
      legend style={draw=none, font=\scriptsize, at={(0.02,0.98)}, anchor=north west},
      legend cell align=left,
    ]
      \addplot[
        only marks,
        mark=*,
        mark size=1.6pt,
        color=black!35
      ] table[x=params, y expr=100-\thisrow{wer}] {\openasravgall};
      \addlegendentry{Other models}

      \addplot[
        no marks,
        line width=0.8pt,
        color=msCyan
      ] table[x=params, y=acc] {\moonshineFrontier};

      \addplot[
        no marks,
        line width=0.8pt,
        color=msOrange
      ] table[x=params, y=acc] {\openaiFrontier};

      \addplot[
        no marks,
        line width=0.8pt,
        color=msGreen
      ] table[x=params, y=acc] {\nvidiaFrontier};

      \addplot[
        only marks,
        mark=*,
        mark size=2.6pt,
        color=msCyan,
        nodes near coords,
        point meta=explicit symbolic,
        every node near coord/.append style={font=\scriptsize, yshift=2pt, anchor=south, text=msCyan}
      ] table[x=params, y expr=100-\thisrow{wer}, meta=label] {\moonshinevTwo};
      \addlegendentry{Moonshine v2}

      \addplot[
        only marks,
        mark=triangle*,
        mark size=2.2pt,
        color=msOrange
      ] table[x=params, y expr=100-\thisrow{wer}] {\openaiModels};
      \addlegendentry{OpenAI}

      \addplot[
        only marks,
        mark=square*,
        mark size=2.2pt,
        color=msGreen
      ] table[x=params, y expr=100-\thisrow{wer}] {\nvidiaModels};
      \addlegendentry{NVIDIA}

      \addplot[
        only marks,
        mark=*,
        mark size=2.8pt,
        color=msGreen,
        nodes near coords,
        point meta=explicit symbolic,
        every node near coord/.append style={
          font=\scriptsize,
          anchor=west,
          xshift=2pt,
          text=msGreen
        }
      ] coordinates {(2500,94.37) [nvidia/canary-qwen-2.5b]};

      \addplot[
        only marks,
        mark=*,
        mark size=2.8pt,
        color=msGreen,
        nodes near coords,
        point meta=explicit symbolic,
        every node near coord/.append style={
          font=\scriptsize,
          anchor=west,
          xshift=5pt,
          yshift=1pt,
          text=msGreen
        }
      ] coordinates {(600,93.95) [nvidia/parakeet-tdt-0.6b-v2]};

      \addplot[
        only marks,
        mark=*,
        mark size=2.8pt,
        color=msOrange,
        nodes near coords,
        point meta=explicit symbolic,
        every node near coord/.append style={
          font=\scriptsize,
          anchor=west,
          xshift=2pt,
          text=msOrange
        }
      ] coordinates {(1550,92.56) [openai/whisper-large-v3]};

      \addplot[
        only marks,
        mark=*,
        mark size=2.8pt,
        color=msOrange,
        nodes near coords,
        point meta=explicit symbolic,
        every node near coord/.append style={
          font=\scriptsize,
          anchor=north east,
          xshift=-2pt,
          yshift=-1pt,
          text=msOrange
        }
      ] coordinates {(809,92.17) [openai/whisper-large-v3-turbo]};
    \end{axis}
  \end{tikzpicture}
  \vspace{-16pt}
  \caption{Accuracy vs. parameter count on Open ASR leaderboard averages.}
  \label{fig:open-asr-avg-scatter}
\end{figure*}
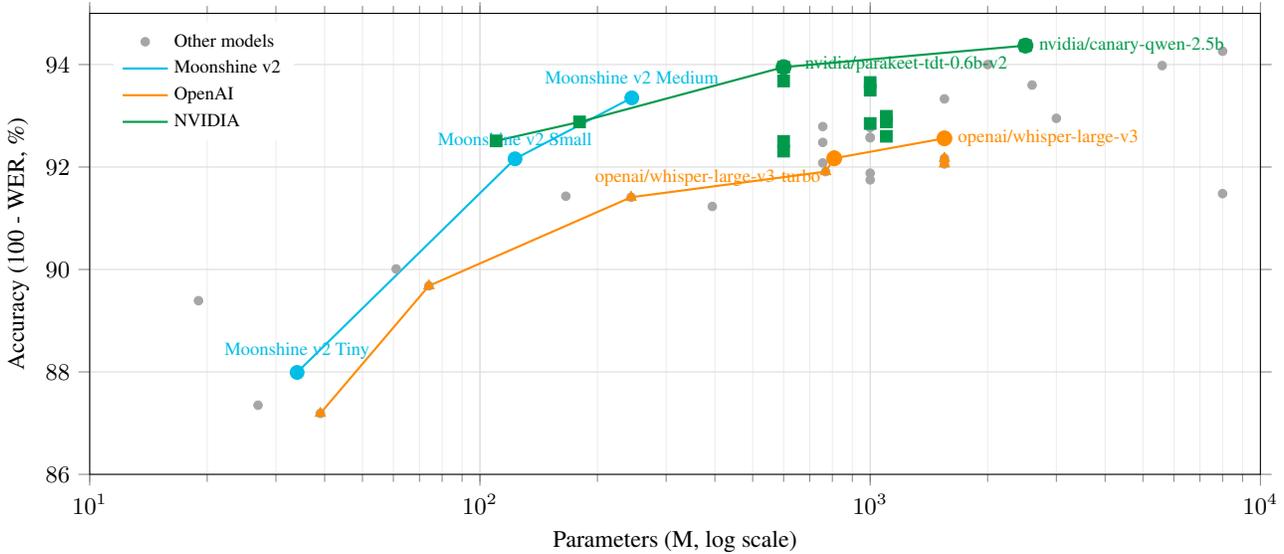

Table~\ref{tab:open-asr-results} reports WERs for individual datasets. We include it
for completeness, but the more informative view is the accuracy--parameter
tradeoff in Figure~\ref{fig:open-asr-avg-scatter}. That plot shows Pareto
frontiers in parameter count versus accuracy. The NVIDIA and Moonshine models
lie on a similar frontier and sit above OpenAI's. Moonshine fills the lower end
of the frontier (in parameter count), which is precisely the region we target:
efficient ASR models for 0.1--1 TOPs and memory-constrained edge processors
(e.g., sub-1~GB). NVIDIA's models, by contrast, are optimized for GPUs with tens
of GB of memory and multi-PFLOP compute.

\begin{figure}[t]
    \centering
    \includegraphics[width=\columnwidth]{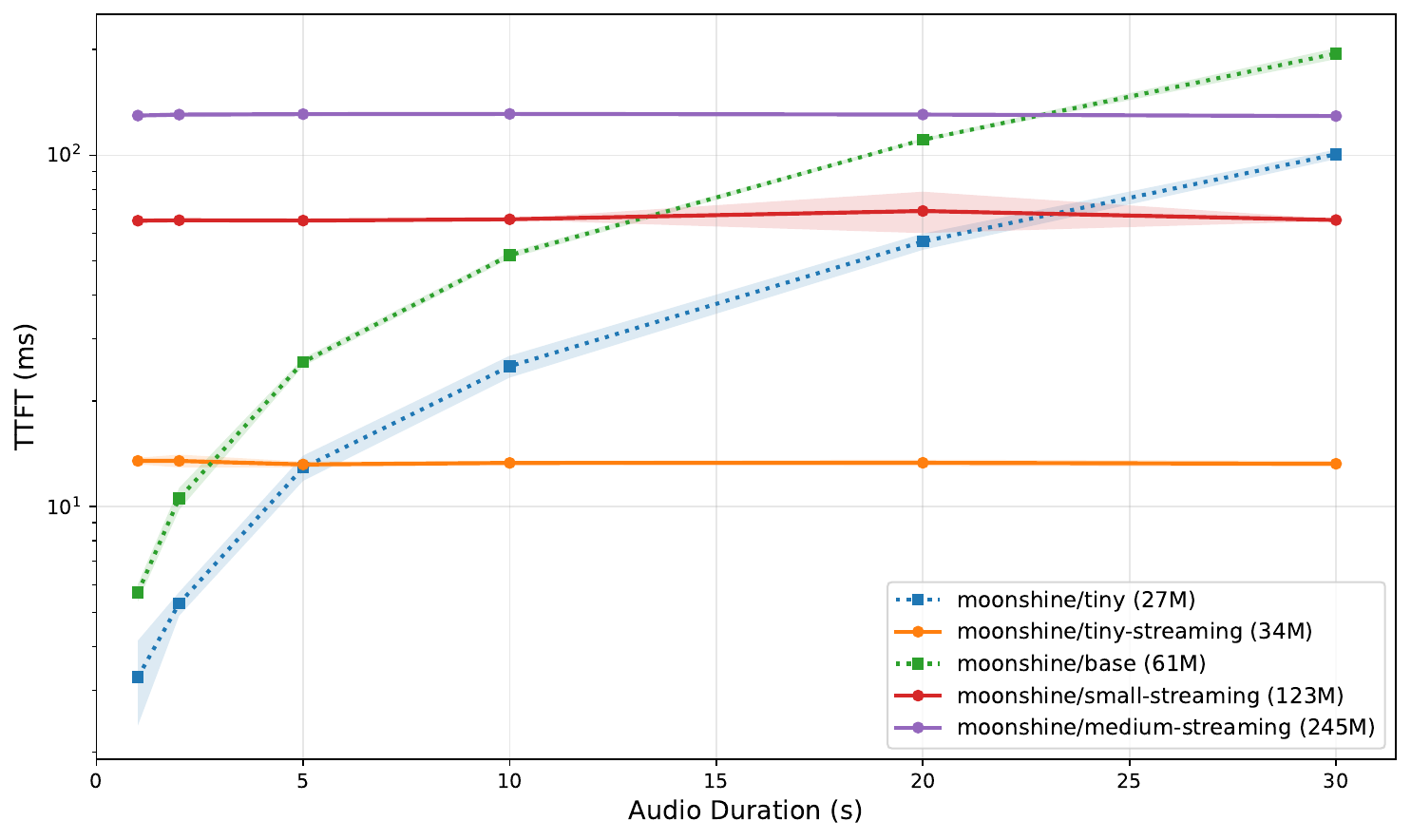}
    \caption{Time to first token (TTFT) versus input audio duration for Moonshine and Moonshine v2 models. The original Moonshine models use a full-attention encoder, which results in TTFT latency that grows with input audio duration. Sliding window attention in the Moonshine v2 encoder results in a fixed encoding latency, regardless of audio duration.}
    \label{fig:ttft-vs-moonshine}
\end{figure}

Figure~\ref{fig:ttft-vs-moonshine} shows the differences in TTFT between full attention and sliding-window attention encoders by comparing the original Moonshine models with Moonshine v2. For longer utterances, even the largest Moonshine v2 achieves lower TTFT to the smaller Moonshine v1 models.

\begin{table}[t]
  \centering
  \small
  \begin{tabular}{l r r}
  \toprule
  Model & Latency (ms) & Compute Load (\%) \\
  \midrule
  \rowcolor{black!10} Moonshine Tiny & 27 & 5.91 \\
  \rowcolor{black!10} Moonshine Base & 44 & 7.34 \\
  \rowcolor{black!10} Moonshine v2 Tiny & 50 & 8.03 \\
  \rowcolor{black!10} Moonshine v2 Small & 148 & 17.97 \\
  \rowcolor{black!10} Moonshine v2 Medium & 258 & 28.95 \\
  Whisper Tiny & 289 & 8.46 \\
  Whisper Base & 553 & 16.19 \\
  Whisper Small & 1940 & 56.84 \\
  Whisper Large v3 & 11286 & 330.65 \\
  \bottomrule
  \end{tabular}
  \caption{Empirical response latency and compute load comparison for Moonshine and Whisper models. These values reflect the amount of time taken between the end of a speech utterance and the returned transcript in a real-time, live transcription scenario running on an Apple MacBook M3.}
  \label{tab:latency-comparison}
\end{table}

Table~\ref{tab:latency-comparison} compares the response latency between Moonshine, Moonshine v2, and Whisper models. The Moonshine v2 models demonstrate substantially lower latency than comparable Whisper models: Moonshine v2 Tiny achieves 50~ms latency (5.8x faster than Whisper Tiny), Moonshine v2 Small achieves 148~ms (13.1x faster than Whisper Small), and Moonshine v2 Medium achieves 258~ms (43.7x faster than Whisper Large v3), while also requiring less compute load on the same hardware.

\begin{table}[t]
  \centering
  \small
  \setlength{\tabcolsep}{6pt}
  \begin{tabular}{l r r r}
  \toprule
  Dataset & Tiny (34M) & Small (123M) & Med. (245M) \\
  \midrule
  AMI & 19.03 & 12.54 & 10.68 \\
  Earnings-22 & 20.27 & 13.53 & 11.90 \\
  GigaSpeech & 13.90 & 10.41 & 9.46 \\
  Libri (clean) & 4.49 & 2.49 & 2.08 \\
  Libri (other) & 12.09 & 6.78 & 5.00 \\
  SPGISpeech & 6.16 & 3.19 & 2.58 \\
  TED-LIUM & 6.12 & 3.77 & 2.99 \\
  VoxPopuli & 14.02 & 9.98 & 8.54 \\
  \midrule
  \textbf{Average} & \textbf{12.01} & \textbf{7.84} & \textbf{6.65} \\
  \bottomrule
  \end{tabular}
  \caption{WER (\%) for Moonshine v2 on Open ASR benchmarks.}
  \label{tab:open-asr-results}
\end{table}

\section{Discussion \& Conclusion}
\label{sec:conclusion}

While our ergodic streaming encoder enables bounded, low-latency TTFT, Moonshine v2 still employs a full-attention autoregressive decoder. This means that once the encoder begins emitting features, the decoder must generate tokens one-by-one through a serial loop. For very long transcripts, this sequential generation can add latency to the full output, even though the first tokens appear quickly. Future work could explore monotonic alignment models or streaming-friendly decoding strategies that further reduce end-to-end latency. Additionally, our current models focus exclusively on English ASR. However, the architectural principles of ergodic streaming encoders with sliding-window attention generalize naturally to other languages. Building on our prior work with specialized, language-specific models~\citep{king2025flavors}, we plan to train Moonshine v2 variants for additional languages, enabling low-latency, on-device speech recognition across diverse linguistic contexts.

We introduced Moonshine v2, a family of streaming ASR models designed for latency-critical, on-device applications. By replacing full-attention encoders with ergodic streaming encoders that use sliding-window self-attention, we achieve bounded TTFT independent of utterance length while maintaining strong transcription accuracy. Our models achieve state-of-the-art results on standard benchmarks, matching the performance of models 6x their size while running significantly faster. These results demonstrate that carefully designed local attention can rival the accuracy of global attention at a fraction of the computational cost, making real-time, interactive speech interfaces practical on resource-constrained edge devices.

\bibliographystyle{icml2022}
\bibliography{paper}

\end{document}